# Two Stages for Visual Object Tracking

**Fei Chen*, Fuhan Zhang and Xiaodong Wang**

College of Computer, National University of Defense Technology, Changsha 410073, China.
*Corresponding Author Email: chenfei14@nudt.edu.cn

**Abstract.** Siamese-based trackers have achieved promising performance on visual object tracking tasks. Most existing Siamese-based trackers contain two separate branches for tracking, including classification branch and bounding box regression branch. In addition, image segmentation provides an alternative way to obtain the more accurate target region. In this paper, we propose a novel tracker with two-stages: detection and segmentation. The detection stage is capable of locating the target by Siamese networks. Then more accurate tracking results are obtained by segmentation module given the coarse state estimation in the first stage. We conduct experiments on four benchmarks. Our approach achieves state-of-the-art results, with the EAO of 52.6% on VOT2016, 51.3% on VOT2018, and 39.0% on VOT2019 datasets, respectively. And our tracker can obtain competitive performance on GOT-10k dataset.

## 1. Introduction

Visual object tracking is a challenging problem in computer vision, which aims to track a target at each frame in a video sequence with the initialization on the first frame. The tracking process is influenced by many challenging scenarios, such as appearance changes, occlusion, illumination, deformation, motion blur, viewpoint change, and fast moving so on. Recently, deep learning has significantly promoted the development of visual object tracking, which obtained fast tracking speed and accurate tracking results, benefiting from the powerful deep features and bounding box regression technologies. While accurate target localization, scale and aspect ratio estimation are still the challenging problem in visual object tracking. In recent years, cross-correlation play an important role which converts the spatial correlation operation into Fourier domain to speeds up the detection process [1]. Afterwards, CFNet [2] firstly integrated the correlation into the convolutional neural networks as a layer and can be trained in an end-to-end manner. Recently, the Siamese-based tracker like SiamRPN [3] equip the Siamese backbone networks with the Region Proposal Networks (RPN) to predict accurate bounding boxes. To enhance the discriminative ability, Ocean [4] adds a global object aware module after the conventional local classifier module within the anchor-free framework. The object aware module aligns the features with the predicted region for each location, which enhance the capability of the classifier to distinguish the target and background. In addition to bounding box regression in most existing deep trackers, segmentation based visual trackers [6] introduce the segmentation network to obtain the rotated bounding boxes, aiming to match well with the datasets like VOT-2016 [7], VOT-2018 [8], and VOT-2019 [9]. While the limitation in SiamMask [5] is that the segmentation branch coupled with two branches such as classification and bounding box lead to the difficulty to train the whole networks. Although applying the deep convolutional neural network (e.g., ResNet50) as backbone to extract features, the robustness still has much room need to be improved, even compared with the trackers based on correlation filters [10],[11]. To address the above limitations, we proposed a two stages based tracker that embrace both the Siamese network and segmentation. Firstly, the outputs of Siamese network provide the coarse localization and scale and ratio estimation of the target. Then, the segmentation network is designed to predict the segmentation mask between target and background. Finally, the rotated bounding box is generated based on the predicted mask. The experimental results illustrated that our two stage tracking-by-segmentation approach outperforms most top trackers on the popular benchmarks such as [7], [8], [9], and [17].

## 2. Related Work

In this section, we review the related tracking methods in details, including CF based trackers, Siamese trackers, and Segmentation based trackers.

### 2.1 Correlation Filter Based Trackers

During the last few years, trackers based on correlation filters (CFs), which was first proposed by Bolme et al. [1], have significantly promoted the tracking performance. Danelljan et al. [12] adds the spatial regularization for penalizing the coefficients of correlation filters based on the spatial location. Li et al. [10] introduce both the spatial and temporal regularizations into the learning. In ASRCF [13], the spatial regularization weight is also learned adaptively like the correlation filter during tracking. To take the advantage of deep features, Valmadre et al. [2] implemented the correlation filters with a layer in a Siamese network and performed tracking by cross-correlation between the template filter and search image. Huang et al. [23] applied proposals for scale and aspect ratio adaptability. Chen et al. [26] used depth information for the correlation filter learning.

### 2.2. Siamese Trackers

Recently, trackers based on Siamese networks have expressed promising performance in visual object tracking, benefiting from their end-to-end framework in training and tracking phases. Bertinetto et al. [14] proposed fully convolutional Siamese network for extracting feature maps of exemplar and search images. The simple cross-correlation layer enables the tracker operates at beyond real-time tracking speed. SiamRPN [3] utilized the Region Proposal Networks to generate prior defined anchors to predict more accurate bounding boxes. DaSiamRPN [15] improved the discriminative ability by the collected hard negative samples in offline training phase and proposed a distractor-aware module to continuously discover hard negative samples (distractors) used for online learning the templates. Afterwards, SiamRPN++ [21] further took the advantages of deeper networks such as ResNet-50 as their backbone and proposed a layer of depth wise cross-correlation for computing the similarity map based on the features of template and search patches, following by a two separate branches for bounding box regression and classification.

### 2.3. Segmentation Based Trackers

Another kind of trackers are that both predict bounding boxes as well as the class-agnostic binary segmentation. In SiamMask [5], Wang et al. proposed an novel framework for producing bounding boxes and the class-agnostic object segmentation masks [16] at the same time. The whole network includes classification branch, box generation branch, and segmentation branch, which can be trained together in offline phase. In D3S [6], they proposed a single-shot discriminative segmentation tracker, which combined geometrically invariant model (GIM) and geometrically constrained Euclidean model (GEM). The geometrically invariant model takes the deep feature vectors of target and background as inputs to produce foreground and background similarity. While the GEM applies the DCF to generate confidence map of target. We found that the performance of the tracking depends heavily on the target localization by GEM model.

## 3. Revisiting the SiamRPN++

SiamRPN++ introduced a depth-wise cross-correlation layer in the RPN module which requires fewer parameters than up-channel cross-correlation in SiamRPN [3]. The template patch **z** is usually given by the bounding box in the first frame of the sequence. For a following frame **x** in the sequence, the goal of SiamRPN++ is to find the best match between the template patch z and search patch x by cross-correlation as:

$$P = \phi(z) \star \phi(x) \quad (1)$$

Where $\phi(z)$ and $\phi(x)$ are the output feature maps of Siamese subnetworks. The template branch and search branch share parameters in backbone network. SiamRPN++ reduce the influence of center bias by a sampling strategy of random translation in offline training. Compared to SiamRPN, the depthwise correlation module can reduce the computation and enables SiamRPN++ to apply more deeper neural networks. In addition, multi-layer aggregation strategy is adopted to enhance the discriminative ability of tracker. To equip deep network like ResNet-50 into the tracking framework, the large stride of 32 pixels in original ResNet-50 has been decreased to 8 pixels to increase the resolution of feature maps of high layers.

## 4. Our Approach

In this section, we present a two-stage tracking framework to fully take advantages of bounding boxes regression of Siamese network and the class-agnostic segmentation module of D3S. As present in Figure 1, our approach includes a backbone network, a detection module, and a segmentation model. The backbone network is utilized for extract feature maps of the template and search patches. The detection module contains two different branches for classification and bounding box regression. The classification branch is utilized to find the target location. The regression branch is aims to produce the bounding box for each of the location.

To reduce computation and adapt to shape variations, we apply an anchor-free detector for target localization. For a feature map $f_i \in R^{H \times W}$, we have $H \times W$ candidate locations for target object. For a ground-truth bounding box $B^i = \{x_0, y_0, x_1, y_1\}$, where $(x_0, y_0)$, $(x_1, y_1)$ are the top-left corner and bottom-right corner, respectively.

Let $(x, y)$ denotes a location in $f_i$, we set it as a positive sample if it falls into any ground-truth bounding boxes $B^i$. The regression target is a 4D real vector $T = \{t^*, l^*, b^*, r^*\}$, where $t^*$, $l^*$, $b^*$, and $r^*$ are the distance from four sides of the bounding box such as top, left, bottom, and right to the target center $(x, y)$. Then the training objective for the target center $(x, y)$ can be computed as:

$$l^* = x - x_0, r^* = x_1 - x, t^* = y - y_0, b^* = y_1 - y \quad (2)$$

At the inference stage, for a location $(c_x, c_y)$, the predicted bounding box can be computed by:

$$x_0 = c_x - l_p^*, y_0 = c_y - t_p^*, x_1 = c_x + r_p^*, x_2 = c_y + b_p^* \quad (3)$$

where $l_p^*$, $t_p^*$, $r_p^*$, and $b_p^*$ are the outputs of bounding box regression branch. Then the final target state is obtained by choosing the position with highest confidence score and the corresponding bounding box.

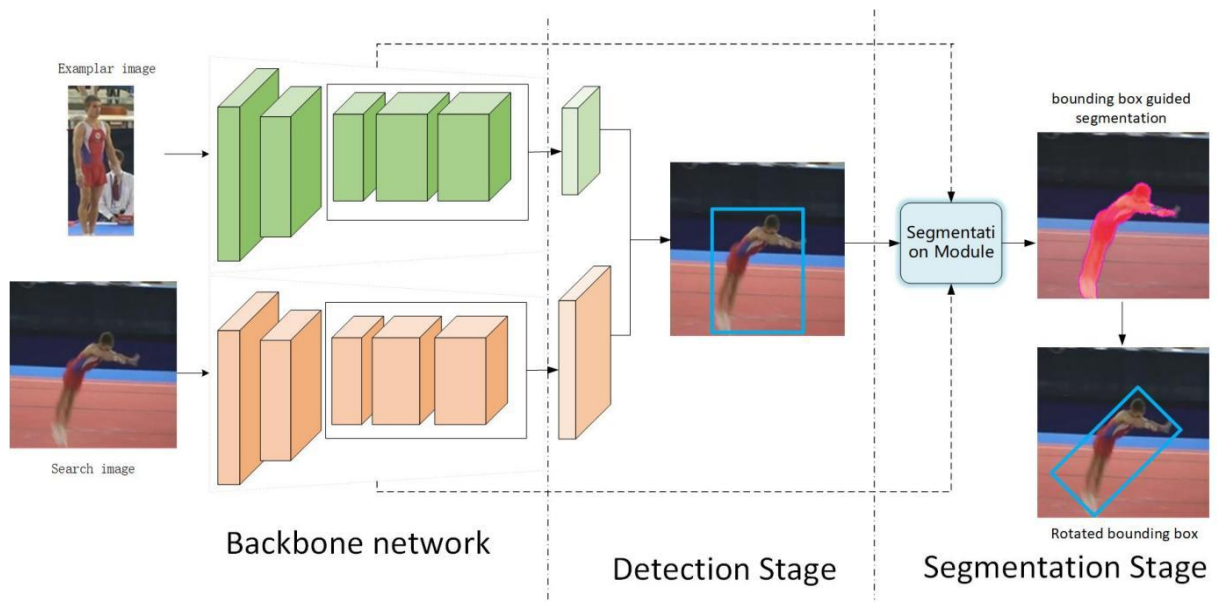

**Figure 1.** Illustration of our tracking approach. The backbone network is utilized to extract feature representations for exemplar patch and search patch. Then the two branches in the detection module are used to predict the location and bounding box of target object. Finally, the target state and deep features from backbone network are feed into segmentation module to estimate the accurate rotated bounding box

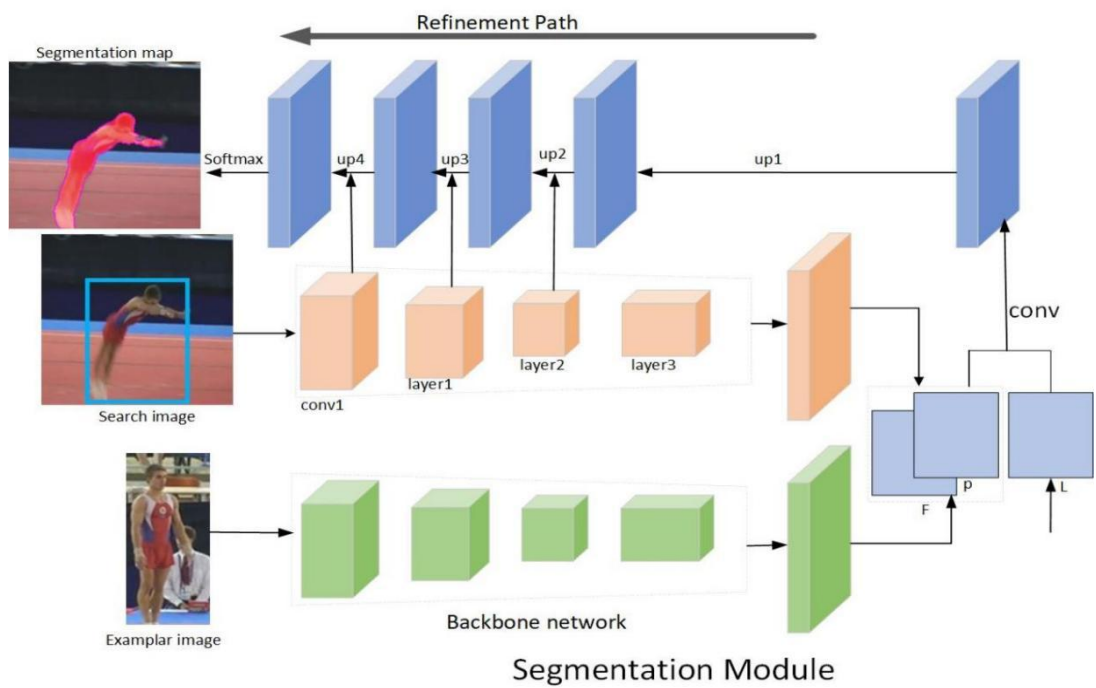

**Figure 2.** The segmentation module in our tracker. In the refinement module, feature maps of conv1, layer1, and layer2 from ResNet-50 are added to last three sampling modules, respectively

When the state of the target was available in the detection stage, we then activate the segmentation module which takes the template patch, search patch, and detected bounding box as inputs. As the Figure 2 shows. We borrow the main framework from D3S [6], while we utilized the ConvTransposed2D module to upsample the feature maps from high layer to be the same with low layer. For a upsampling layer, the feature maps from the backbone are added to enrich its information. The final upscaling module combined with a softmax function are utilized to generate the segmentation confidence map. Then we binarize the output of probability map with a threshold 0.5 to produce the final mask.

## 5. Experiments

We evaluate our tracker on four challenging datasets, including VOT-2016 [7], VOT-2018 [8], VOT-2019 [9], and GOT-10k [17] datasets. The performance in VOT datasets is evaluated by accuracy and robustness. The accuracy computes the average overlap among the successfully tracked frames. The robustness reflects the failure times of the tracker. A failure indicates that the there is a zero overlap between the ground truth and predicted bounding box. The EAO integrates the accuracy and robustness, which aims to reduce the variance and bias come from AverageOverlap measure because of different sequence length.

*5.1 Implementation Details*

For the detector in first stage, we apply the online tracker of Ocean [4] as our baseline. The ResNet-50 is utilized as the backbone for both the detector and segmentation modules. The segmentation networks are pre-trained on the Youtube-VOS [19] dataset. Our approach is implemented using Python 3.7, PyTorch 1.1.0 with Intel(R) i5-7500 3.4 GHz CPU, 64 RAM, and NVIDIA GTX 1080Ti.

*5.2 Comparison with the State-of-the-art*

In this section, we evaluate our approach with several top trackers on four datasets.

VOT-2016 [7]: There are 60 sequences in this dataset with more accurate ground-truth bounding boxes than VOT-2015. Each frame in this dataset is labelled with five attributes such as illumination variations, occlusion, motion changes, camera motion, and size variations. As shown in Table 1, we compare our approach with several outstanding trackers include ECO [20], SiamRPN++ [21], SiamBAN [22], DaSiamRPN [15], D3S [6], ATOM [24], and DiMP [25]. Compared with D3S, our approach achieves similar accuracy, but obtains a significant gain of 3.3% over D3S in terms of EAO and accuracy increases by 0.05%. Compared to above trackers, our approach obtains highest scores for accuracy and EAO.

**Table 1.** Tracking results on VOT-2016 dataset, where A denotes the accuracy and Ro denotes the robustness

|      | ECO   | ATOM  | DiMP  | SiamRPN++ | DaSiamRPN | SiamBAN | D3S   | **Ours** |
|------|-------|-------|-------|-----------|-----------|---------|-------|----------|
| EAO↑ | 0.374 | 0.424 | 0.479 | 0.478     | 0.401     | 0.505   | 0.493 | **0.526** |
| A↑   | 0.54  | 0.617 | 0.624 | 0.637     | 0.609     | 0.632   | 0.66  | **0.671** |
| Ro↓  | 0.2   | 0.189 | 0.135 | 0.177     | 0.224     | 0.149   | 0.131 | **0.131** |

VOT-2018 [8]: All sequences in VOT-2018 maintains the same sequences with VOT2016, in which 10 new challenging sequence are added to replace the least challenging ones. As shown in Table 2.Tracking results on VOT-2018 dataset., our tracker achieves the EAO of 51.3% and accuracy of 66.2%, respectively. Our approach achieves a gain of 2.4% over D3S and OceanOn. Our tracker ranks second in terms of robustness.

**Table 2.** Tracking results on VOT-2018 dataset.

|      | ATOM  | DiMP  | SiamRPN++ | DaSiamRPN | SiamBAN | D3S   | OceanOn | **Ours** |
|------|-------|-------|-----------|-----------|---------|-------|---------|----------|
| EAO↑ | 0.401 | 0.44  | 0.415     | 0.326     | 0.447   | 0.489 | 0.489   | **0.513** |
| A↑   | 0.59  | 0.597 | 0.6       | 0.57      | 0.59    | 0.64  | 0.592   | **0.662** |
| Ro↓  | 0.201 | 0.152 | 0.234     | 0.337     | 0.178   | 0.15  | 0.117   | **0.15**  |

VOT-2019 [9]: In this dataset, 12 new sequences selected from a subset dataset from GOT-10k. And the 12 least difficult sequences of VOT2018 are replaced by the new selected sequences. Meanwhile, VOT-2019 provides the annotated segmentation masks and rotated bounding boxes for tracking targets. Compared with previous datasets, VOT-2019 poses new challenging for trackers. As shown in Table 3, Our tracker obtains a gain of 4.2% and 4% over D3S and OceanOn in terms of EAO, respectively. Compared to D3S, the accuracy increases to 65.8% and the failure rates deceases to 27.6%. This illustrates that our tracker can obtain more accurate bounding boxes with good robustness.

Table 3. Tracking results on VOT-2019 dataset.

|   | SiamMASK | ATOM | DiMP | SiamRPN++ | SiamBAN | D3S | OceanOn | **Ours** |
|---|---|---|---|---|---|---|---|---|
| EAO↑ | 0.287 | 0.301 | 0.379 | 0.292 | 0.322 | 0.348 | 0.350 | **0.390** |
| A↑ | 0.594 | 0.603 | 0.594 | 0.580 | 0.596 | 0.646 | 0.594 | **0.658** |
| Ro↓ | 0.461 | 0.411 | 0.278 | 0.446 | 0.396 | 0.331 | 0.316 | **0.276** |

GOT-10k [17]: This is a large-scale tracking dataset. The train set consists of 10,000 sequences and test set contains 180 sequences. There is no overlap between these two subsets. Two metrics include Average Overlap (AO) and Success Rate(SR) are used to measure the trackers. Average Overlap computes the average overlaps between predicted bounding boxes and its ground truth. Success Rate denotes the ratio of successfully tracked frames in which the overlaps are larger than 50% to the total frames of a sequence. The Table 4 presents the overall performance of our tracker and other seven trackers. Our approach can achieve competitive performance compared to others. Compared to D3S, our tracker obtains gains of 0.7% and 2.1% in terms of AO and $SR_{0.5}$, respectively. Our method has a comparable AO score with OceanOn, which obtains the highest AO score.

Table 4. Tracking results on GOT-10k dataset. The $SR_{0.5}$ is SR score in which the overlap is larger than 50%, and $SR_{0.75}$ denotes the SR score in which the overlap is larger than 0.75%

|   | SiamMASK | ATOM | DiMP | SiamFc | SiamFCv2 | D3S | OceanOn | Ours |
|---|---|---|---|---|---|---|---|---|
| AO↑ | 0.514 | 0.556 | 0.611 | 0.348 | 0.374 | 0.597 | 0.611 | 0.604 |
| $SR_{0.5}$ ↑ | 0.587 | 0.635 | 0.717 | 0.353 | 0.404 | 0.676 | 0.721 | 0.697 |
| $SR_{0.75}$ ↑ | 0.366 | 0.402 | 0.492 | 0.098 | 0.141 | 0.462 | 0.493 | 0.465 |

## 6. Conclusion

In this paper, a two stage tracker was proposed, aiming to track the target by taking the advantages of object detection and image segmentation. We evaluate our approach on four public tracking benchmarks such as VOT-2016, VOT-2018, VOT-2019, and GOT-10k. The tracking results on these datasets illustrated that our approach can achieve competitive performance in terms of accuracy and robustness.

## 7. Acknowledgments

This work was supported by Science and Technology Foundation of State Key Laboratory under Grant 6142110180405.